\documentclass[11pt,letter]{article}

\usepackage[utf8]{inputenc}
\usepackage[T1]{fontenc}
\usepackage{amsmath,amssymb,amsfonts}
\usepackage{graphicx}
\usepackage[margin=1in]{geometry}
\usepackage{setspace}
\usepackage{hyperref}
\usepackage{cite}
\usepackage{siunitx}
\usepackage{lineno}
\usepackage{subcaption}
\graphicspath{{figures/}}

\hypersetup{
    colorlinks=true,
    linkcolor=blue,
    citecolor=blue,
    urlcolor=blue
}

\title{An Open-Source Two-Stage Computer Vision Pipeline for Fine-Grained Vehicle Classification using Vision Transformers}

\author{
    Gandhimathi Padmanaban\textsuperscript{1}, Fred Feng\textsuperscript{1,*}
}

\date{}

\begin{document}


\maketitle

\begin{center}
\small
\textsuperscript{1}Department of Industrial Manufacturing and Systems Engineering, University of Michigan-Dearborn, Michigan, USA, 48128\\
\textsuperscript{*}Corresponding author: fredfeng@umich.edu
\end{center}

\begin{abstract}
Vehicle body type is a significant determinant of cyclist injury severity in overtaking
crashes, yet automated tools capable of classifying vehicles into injury-risk-relevant
categories from naturalistic roadway video do not exist in the open literature. Standard
object detection benchmarks provide only coarse vehicle labels (car, truck, bus,
motorcycle), while existing fine-grained recognition systems are trained on controlled
imagery and have not been evaluated for deployment robustness across recording sites.
This paper presents an open-source two-stage computer vision pipeline that combines a
pre-trained RT-DETR detector for coarse vehicle localization with a fine-tuned Vision
Transformer (ViT-Base/16) for six-category body-type classification: passenger car, SUV,
pickup truck, minivan, large van, and commercial truck. A confidence-based abstention
mechanism withholds Stage~2 predictions when the softmax output falls below a
class-specific threshold of 0.60, producing \textit{unknown} labels rather than silent
misclassifications. The pipeline was evaluated on 3,805 annotated overtaking events from
a bicycle-lane corridor in Ann Arbor, Michigan (in-distribution), achieving an overall
accuracy of 0.94 with per-class F1 scores ranging from 0.91 (minivan) to 0.97 (SUV).
On an independent out-of-distribution evaluation of 311 events from an open cycling dataset collected using an
instrumented research bicycle without retraining, the pipeline achieved an
overall accuracy of 0.89. Three of the four well-represented categories maintained F1
scores at or above 0.90 under this domain shift. The largest cross-domain degradation
was observed for minivan (F1 = 0.72), driven primarily by a rise in the abstention rate
from 2.4\% to 25.0\% rather than by active misclassification, consistent with the
abstention mechanism propagating genuine model uncertainty rather than producing silent
errors. The full pipeline, including inference scripts, training code, evaluation
utilities, and fine-tuned model weights, is released as open-source software to support
reproducibility and reuse across roadside video archives and cycling safety research
applications.

\vspace{3em}

\end{abstract}

\newpage

\section{Introduction}
\label{sec:intro}

\subsection{Motivation: Vehicle Type and Cyclist Safety}
\label{sec:intro_motivation}

Cycling fatalities in the United States have increased substantially over the past two decades, with pedalcyclist deaths rising from 786 in 2010 to 1,105 in 2022, representing 2.6\% of all traffic fatalities~\cite{ntsb2019bicyclist}. While multiple factors contribute to this trend, vehicle composition plays an increasingly prominent role. Research has established that SUVs and light truck vehicles (LTVs) cause more severe injuries to pedestrians and cyclists than passenger cars in equivalent crash events~\cite{robinson2025suv,monfort2023bicyclist}. Specifically, the odds of fatal injury increase by 44\% when a pedestrian or cyclist is struck by an SUV compared to a passenger car, rising to 82\% for children~\cite{robinson2025suv}. The mechanism is well established: the elevated front-end geometry of SUVs and pickup trucks strikes cyclists above their center of gravity, causing knockdown and secondary run-over injuries rather than the forward-vault pattern typical of passenger car impacts~\cite{monfort2023bicyclist}. Taller front-end vehicles are also disproportionately associated with head injuries, which account for the majority of cyclist fatalities~\cite{beck2015cyclist}.

These differential injury risks motivate disaggregated exposure analyses: understanding not merely how many vehicles overtake a cyclist, but \textit{which types} do so. Passing-distance legislation, now enacted in over 35 U.S. states with minimum clearances of three to four feet, creates an additional policy imperative. Research has demonstrated that such laws significantly increase driver overtaking
distances~\cite{feizi2021passinglaw}, yet most measurements of cyclist passing proximity
have relied on instrumented bicycles in naturalistic studies~\cite{walker2007drivers}
rather than fixed roadside cameras capable of simultaneous vehicle-type annotation. Video data offer a richer source for exposure measurement, enabling facility-level and
vehicle-level disaggregation not possible from GPS or self-report data
alone~\cite{hamann2017beyondgps,liu2016egocentric}. Existing methods for generating vehicle-type exposure data either rely on manual annotation (which does not scale to large video archives) or on traffic counting technologies that produce only coarse vehicle classifications insufficient to distinguish, for example, a passenger SUV from a commercial van.

\subsection{Computer Vision for Traffic Monitoring and Vehicle Classification}
\label{sec:intro_cv}

Computer vision has matured into a productive technology for traffic monitoring. Roadside camera systems have been applied to vehicle detection, tracking, speed estimation, and behavior analysis across a wide range of deployment contexts~\cite{datondji2016survey, zangenehpour2015automated}. The development of deep convolutional neural networks, catalyzed by the ImageNet benchmark~\cite{russakovsky2015imagenet}, produced object detectors (from Faster R-CNN to the YOLO series) capable of identifying vehicles in real time. The introduction of Transformer-based end-to-end detectors, beginning with DETR~\cite{carion2020detr} and culminating in RT-DETR~\cite{zhaoDETR}, has further improved accuracy and reduced reliance on hand-designed post-processing components such as non-maximum suppression, making transformer detectors increasingly attractive for deployment on standard roadside hardware.

Vehicle \textit{classification}, the assignment of a detected vehicle to a functional body-type category, is a distinct and more demanding task than detection. Standard object detection benchmarks such as COCO provide only four vehicle categories
(car, truck, bus, motorcycle), a granularity insufficient for most transportation safety analyses.
Infrastructure-based detectors developed for VRU safety contexts have similarly noted
this coarse-category limitation and proposed specialized architectures to improve
detection of cyclists and pedestrians from roadside cameras~\cite{shi2024enhancedvru}. Research on fine-grained vehicle recognition has largely focused on make-and-model identification, treating the Stanford Cars dataset~\cite{stanford_cars} as the canonical benchmark. CNN-based approaches on this benchmark have achieved top-1 validation accuracies above 90\% across the full 196-class make-and-model vocabulary~\cite{corrales2019cnns,wang2020multipath}. More recently, Vision Transformers (ViTs)~\cite{dosovitskiy2021vit} have demonstrated competitive performance on fine-grained visual categorization tasks, leveraging global self-attention to capture long-range spatial dependencies between vehicle parts that local convolutional filters may miss~\cite{naz2025cvit}.

Despite these advances, direct application of make-and-model recognition to roadway safety problems is limited for two reasons. First, the functional body-type categories relevant to injury risk (passenger car, SUV, pickup truck, minivan, large van, commercial truck) cut across hundreds of make-and-model classes and are not directly recoverable from model-level predictions without an additional mapping layer. Second, existing fine-grained recognition systems are almost universally trained and evaluated on clean, frontal, or near-ideal imagery such as manufacturer promotional photographs or controlled acquisition setups; their performance on naturalistic roadway video (with variable viewpoint, motion blur,
partial occlusion, variable lighting, and domain shift across recording sites) has been rarely assessed~\cite{liu2023transfer, fu2017automatic}.

\subsection{Two-Stage Detection-Classification Architectures}
\label{sec:intro_twostage}

The coarse-to-fine paradigm, in which a system detects objects first and then classifies them at higher specificity, has a well-established precedent in computer vision. In the autonomous driving perception literature, two-stage 3D object detectors such as PV-RCNN decouple region proposal from attribute refinement to achieve higher accuracy than single-stage end-to-end networks on complex scenes~\cite{shi2020pvrcnn}. Analogous decompositions have been applied in fine-grained vehicle recognition~\cite{fang2017fgvc}, where a coarse detection stage reduces the search space and filters background before a more discriminative second-stage network operates on cropped regions. This modular architecture offers a practical deployment advantage: the detection backbone can be a general-purpose pre-trained model applied without task-specific fine-tuning, while the fine-grained classifier is trained on a smaller, domain-targeted dataset. The approach also enables selective invocation of the computationally heavier classifier (only on detections that require fine-grained label assignment), which reduces inference cost relative to applying a full fine-grained model to every pixel of the full-resolution frame.

For the specific problem of vehicle taxonomy in transportation research, the closest precedent is the system of Almutairi et al.~\cite{almutairi2022truck}, which deployed a CNN on highway camera footage to classify tractor-trailer configurations for freight flow modeling. That system addressed a specialized and relatively constrained sub-problem (distinguishing truck body and trailer configurations), operated on a single camera deployment, and did not evaluate cross-site generalization. Recent work has begun exploring fixed-site video tools for cyclist overtaking safety
analysis~\cite{toulouse2025overtakingvideo}, yet none incorporate automated vehicle body-type
classification at the specificity required for injury-risk-stratified exposure studies.
More generally, the gap between freight-oriented truck taxonomy systems and the broader
passenger-vehicle body-type vocabulary required for cyclist exposure research has not been
addressed in the literature.

\subsection{Research Gap}
\label{sec:intro_gap}

A review of existing work reveals three compounding gaps that the present study is designed to address:

\begin{enumerate}
    \item \textbf{Vocabulary mismatch.} Existing fine-grained vehicle recognition benchmarks and deployed systems target either coarse COCO-level categories (insufficient for safety analysis) or make-and-model labels (not aligned with injury-risk-relevant body-type categories). A system trained and evaluated explicitly on a six-category body-type taxonomy (passenger car, SUV, pickup truck, minivan, large van, commercial truck) does not exist in the open literature.

    \item \textbf{Deployment gap.} Prior systems are predominantly trained and evaluated on clean, controlled-acquisition imagery. Naturalistic roadway footage from roadside cameras presents substantially different image characteristics: oblique viewpoints, motion blur at frame edges, partial occlusion by preceding vehicles, variable lighting, and diverse backgrounds. Performance under these conditions, including out-of-distribution generalization across recording sites, has not been systematically reported for body-type classification.

    \item \textbf{Accessibility and reproducibility.} No open-source, end-to-end pipeline integrating real-time detection with fine-grained body-type classification exists that can be applied to existing roadside video archives without custom infrastructure or proprietary data. The research community studying cycling safety, traffic composition, and vulnerable road user (VRU) exposure lacks a shared, reproducible tool for automated vehicle-type annotation.
\end{enumerate}

\subsection{Contributions}
\label{sec:intro_contributions}

This paper addresses these gaps through the following contributions:

\begin{enumerate}
    \item \textbf{A two-stage classification pipeline} combining a pre-trained RT-DETR detector for coarse vehicle localization with a fine-tuned Vision Transformer (ViT-Base/16) for six-category fine-grained body-type classification. The system requires no bounding-box annotation for the detection stage and is deployable on standard roadside video without infrastructure modification.

    \item \textbf{A confidence-based abstention mechanism} that withholds predictions when Stage~2 softmax confidence falls below a class-specific threshold, producing calibrated \textit{unknown} labels rather than silent misclassifications. This mechanism is evaluated alongside standard accuracy metrics, providing a complete picture of the pipeline's operational reliability.

    \item \textbf{A large-scale naturalistic evaluation} on 3,805 annotated overtaking events from a bicycle-lane corridor (in-distribution) and 311 events from an independent dataset collected at separate sites (out-of-distribution), providing the first cross-site generalization assessment for a six-class body-type classification system applied to roadside video.

    \item \textbf{A publicly released open-source implementation} including inference scripts, training code, evaluation utilities, and the fine-tuned ViT model weights, enabling direct replication and extension of all reported results (Section~\ref{sec:availability}).
\end{enumerate}

The remainder of the paper is organized as follows. Section~\ref{sec:methods} describes the study site, training dataset construction, pipeline architecture, and evaluation protocol. Section~\ref{sec:results} reports in-distribution and out-of-distribution classification performance across all six vehicle categories. Section~\ref{sec:discussion} interprets the findings in relation to prior work, addresses limitations, and outlines implications for cycling safety research and future directions. Section~\ref{sec:conclusions} summarizes the principal conclusions.

\section{Method}
\label{sec:methods}

This section describes the study site and data collection approach, the construction of the training dataset, the architecture and training of each pipeline stage, the inference logic applied at deployment, and the evaluation protocol used to assess performance.

\subsection{Pipeline Overview}
\label{sec:pipeline_overview}

\begin{figure}[h]
    \centering
    \includegraphics[]{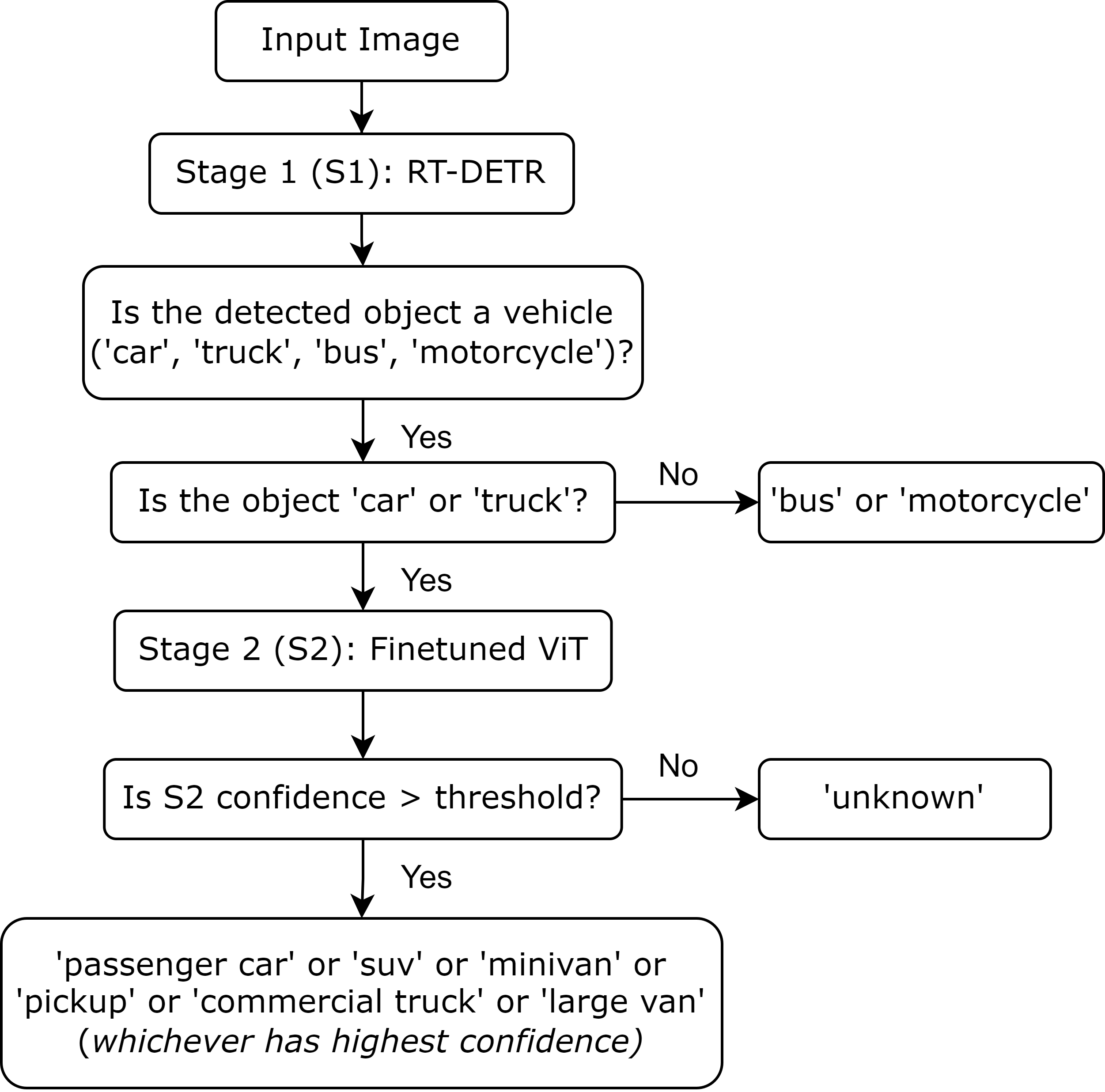}
    \caption{Decision logic of the two-stage classification pipeline. Stage~1 (RT-DETR) detects all vehicles and assigns coarse COCO labels. Detections classified as car or truck are forwarded to Stage~2 (fine-tuned ViT) for fine-grained classification. Bus and motorcycle detections bypass Stage~2 and retain their Stage~1 label. A confidence threshold gates the Stage~2 output; predictions below the threshold are returned as \textit{unknown}.}
    \label{fig:pipeline}
\end{figure}

The proposed system is a sequential two-stage pipeline. In the first stage, a pre-trained real-time object detector scans each video frame and returns axis-aligned bounding boxes for all vehicles present, along with coarse COCO category labels (car, truck, bus, motorcycle). In the second stage, detections belonging to the car or truck categories are cropped from the original frame and passed to a fine-tuned Vision Transformer (ViT), which assigns one of six fine-grained vehicle type labels. Detections classified by the Stage~1 detector as bus or motorcycle are retained with their coarse label without invoking the ViT, as these categories fall outside the scope of the classification task. A confidence-based abstention mechanism allows the Stage~2 classifier to withhold a prediction when the softmax probability of the top class falls below a predefined threshold, rather than emitting a low-confidence label. Figure~\ref{fig:pipeline} shows the complete decision logic of the pipeline. The full implementation, including inference scripts, training code, evaluation utilities, and fine-tuned model weights, is released as open-source software to support reproducibility and reuse (see Section~\ref{sec:availability}).

\subsection{Study Site, Video Data, and Ground-Truth Annotation}
\label{sec:data_collection}

\subsubsection{Field data collection}
Video data were collected at the Ann Arbor N.~Division Street corridor, a shared roadway with an adjacent bicycle lane that serves as the primary in-distribution evaluation site. Data from two independent collection sessions, conducted on October~6, 2022 and October~11, 2022, were used. An additional out-of-distribution evaluation set was drawn from an open cycling dataset collected using a specially instrumented research bicycle equipped with
synchronized lidar, camera, and inertial sensors~\cite{feng2024nsfcyclingsafety}, recorded at locations and on different dates from the Ann Arbor training-domain footage.

Raw video from both datasets was decoded at 10 frames per second using \texttt{ffmpeg}, yielding sequentially numbered JPEG frames. No frame preprocessing (e.g., denoising, contrast enhancement) was applied prior to inference.

\subsubsection{Ground-truth annotation}
For each data collection session, human annotators reviewed the decoded frame sequence and identified the frame number corresponding to the moment an overtaking vehicle passed the camera. Each such event was recorded in a comma-separated file alongside a manually assigned vehicle type label drawn from the set \{passenger car, SUV, pickup truck, minivan, large van, commercial truck\}. Annotation quality was reviewed iteratively; frames where the vehicle was substantially occluded, the type was ambiguous, or the ground-truth label did not match any of the six target categories were excluded from quantitative evaluation. The two Ann Arbor sessions together yielded 3,874 annotated passing events (2,030 on October~6; 1,844 on October~11), of which 3,805 were retained after filtering for the six-class evaluation (see Section~\ref{sec:eval_protocol}). The excluded events correspond to categories outside the six-class vocabulary (bus, motorcycle, other).

\subsection{Training Dataset Construction}
\label{sec:dataset}

Fine-grained vehicle type recognition presents a class-imbalance challenge: passenger cars and SUVs are substantially more common in publicly available labeled image collections than large vans or commercial trucks. To address this, the training set was assembled from three complementary sources.

\subsubsection{Stanford Cars}
The Stanford Cars dataset \cite{stanford_cars}, which provides model-level annotations across 196 make-and-model categories, was mapped to four of the six target categories (passenger car, SUV, pickup truck, minivan) using a rule-based label translator that matched manufacturer and body-type keywords in the original class names. After mapping and excluding ambiguous or out-of-scope categories (e.g., cargo vans, commercial vehicles), 14,478 images were retained across the four classes.

\subsubsection{Web-scraped imagery}
To supplement the three categories with limited representation in Stanford Cars, images of large vans, commercial trucks, and minivans were collected from publicly accessible web sources using a structured image scraper. Downloaded images were manually reviewed for label accuracy and image quality before inclusion; corrupt, duplicate, and ambiguous images were removed.

\subsubsection{Field crops from Ann Arbor footage}
Vehicle crops for all six categories were extracted directly from the Ann Arbor N.~Division frame sequences using the Stage~1 detector described in Section~\ref{sec:stage1}. Detections were matched to annotated passing events to provide in-domain imagery reflecting the specific camera viewpoint, lighting conditions, and occlusion patterns of the deployment environment. This source is particularly important for categories such as large van and commercial truck, which are infrequent even in web-collected datasets.

The three sources were merged into a single flat directory structure organized by class. Table~\ref{tab:dataset_composition} summarizes the per-class and per-source sample counts. The resulting dataset of 16,581 images is heavily skewed: passenger car and SUV together account for 86\% of training images, while commercial truck (0.5\%) and large van (0.6\%) together represent fewer than 200 samples. This imbalance directly motivates the class-balancing strategy described in Section~\ref{sec:stage2}. Notably, commercial truck and large van remain sparsely represented in the evaluation datasets as well (10 and 52 ground-truth events in-distribution; 10 and 9 out-of-distribution), so per-class performance estimates for these categories carry higher uncertainty than those for the four more common classes.

\begin{table}[h]
\centering
\caption{Training dataset composition by class and source ($N = 16{,}581$). Stanford Cars contributes only passenger car, SUV, pickup truck, and minivan. Web-scraped images supply the three under-represented categories. Ann Arbor field crops cover all six classes and provide in-domain imagery.}
\label{tab:dataset_composition}
\begin{tabular}{lrrrrc}
\hline
\textbf{Class} & \textbf{Stanford Cars} & \textbf{Web-scraped} & \textbf{Field crops} & \textbf{Total} & \textbf{\%} \\
\hline
Passenger car     & 9,689 &   0 &   686 & 10,375 & 62.6\% \\
SUV               & 2,854 &   0 & 1,034 &  3,888 & 23.4\% \\
Pickup truck      & 1,519 &   0 &   111 &  1,630 &  9.8\% \\
Minivan           &   416 &  33 &    61 &    510 &  3.1\% \\
Large van         &     0 &  68 &    24 &     92 &  0.6\% \\
Commercial truck  &     0 &  84 &     2 &     86 &  0.5\% \\
\hline
\textbf{Total}    & \textbf{14,478} & \textbf{185} & \textbf{1,918} & \textbf{16,581} & \textbf{100\%} \\
\hline
\end{tabular}
\end{table}

\subsection{Stage 1: Vehicle Detection}
\label{sec:stage1}

The first pipeline stage uses RT-DETR \cite{zhaoDETR}, a transformer-based real-time end-to-end object detector. Specifically, the \texttt{PekingU/rtdetr\_r50vd\_coco\_o365} checkpoint, pre-trained jointly on COCO and Objects365, was used without fine-tuning. This checkpoint was selected because joint pre-training on Objects365 improves recall for less common vehicle categories (e.g., large vans) compared to models trained on COCO alone.

At inference time, the detector is applied to the full-resolution video frame and returns bounding boxes with associated class probabilities for all objects. Only detections belonging to the four COCO vehicle categories (car, truck, bus, motorcycle; class IDs 2, 3, 5, 7) are retained. All other detections are discarded before subsequent processing. A Stage~1 confidence threshold of 0.35 was applied; detections below this threshold are not forwarded to Stage~2. This value was chosen through manual inspection of detection outputs to suppress spurious bounding boxes on background objects while retaining vehicles that are partially out of frame or at lower resolution.

A minimum size gate is applied before invoking the Stage~2 classifier on car or truck detections: bounding boxes whose shorter side is less than 5.5\% of the frame height, or whose area is less than 0.04\% of the total frame area, are excluded from Stage~2 classification. These detections represent vehicles that are too distant for the ViT crop to contain meaningful type-discriminating texture, and retaining their coarse Stage~1 label avoids potentially misleading fine-grained predictions.

\subsection{Stage 2: Fine-Grained Classification with Vision Transformer}
\label{sec:stage2}

\subsubsection{Architecture}
The second stage classifier is based on \texttt{google/vit-base-patch16-224-in21k} \cite{dosovitskiy2021vit}, a Vision Transformer with a patch size of $16 \times 16$ pixels pre-trained on ImageNet-21k. The classification head was replaced with a six-class linear layer and the full model was fine-tuned on the assembled training dataset. Input images were resized to $256 \times 256$ pixels and center-cropped to $224 \times 224$ for inference; at training time, a random crop of $224 \times 224$ was used instead of center crop to introduce spatial variation.

\subsubsection{Training augmentation}
Training augmentations included random horizontal flip, color jitter (brightness, contrast, and saturation factors of 0.3; hue factor of 0.1), and random erasing with probability 0.2. These augmentations were applied on-the-fly during training using standard torchvision transforms. No augmentation was applied during inference.

\subsubsection{Loss function and class balancing}
Training used focal loss \cite{Lin_2017_ICCV} with a focusing parameter $\gamma = 2.0$ and per-class weights derived from inverse class frequency:

\begin{equation}
    w_c = \frac{N}{n_c \cdot C}
    \label{eq:class_weight}
\end{equation}

\noindent where $N$ is the total number of training samples, $n_c$ is the number of samples in class $c$, and $C$ is the number of classes. These weights were normalized so that $\sum_c w_c = C$. In addition, a weighted random sampler was used during training to oversample minority classes within each batch, providing a complementary mechanism to focal loss for handling the imbalance between common and rare vehicle types.

\subsubsection{Optimization}
The model was trained for 30 epochs with the AdamW optimizer at a learning rate of $2 \times 10^{-5}$ and a batch size of 64. Mixed-precision (FP16) training was used when a GPU was available. No validation split was held out; the model was trained on the full assembled dataset, and generalization was assessed using the separately collected field evaluation datasets described in Section~\ref{sec:eval_protocol}. The final checkpoint was saved after completion of the 30th epoch.

\subsection{Classification Confidence and Abstention}
\label{sec:abstention}

A per-class confidence threshold of 0.60 is applied to the Stage~2 softmax output. When the maximum predicted probability falls below this threshold, the pipeline assigns the label \textit{unknown} rather than the top-ranked class. This abstention mechanism is intended to reduce the rate of silent misclassifications in conditions where the image crop is ambiguous (e.g., due to motion blur, extreme viewing angle, or partial occlusion). In the evaluation reported here, \textit{unknown} predictions are treated as incorrect classifications rather than abstentions, so that the reported accuracy figures reflect a worst-case bound. The abstention rate per class is reported separately as a complementary diagnostic.

\subsection{Evaluation Protocol}
\label{sec:eval_protocol}

\subsubsection{Matching and normalization}
Inference outputs and ground-truth labels were joined on the annotated frame number. Ground-truth label strings were normalized to the canonical six-class vocabulary (e.g., ``car'' was mapped to passenger car, ``cargo van'' to large van) before comparison. Frames with no detection in the inference output were excluded from metric computation, as they reflect pipeline failures at the detection stage rather than classification errors.

\subsubsection{Evaluation sets}
Two evaluation conditions were used. The \textit{in-distribution} condition used the two Ann Arbor N.~Division sessions, which share the same recording site and camera setup as the field crops used in training; these sessions provided 3,805 matched events across all six classes. The \textit{out-of-distribution} condition two instrumented bicycle dataset trips, which were collected at different sites and under different conditions from the training data, providing 311 matched events across all six classes after filtering.

Results are reported for all six classes in both conditions. Large van ($n_\text{ID} = 52$, $n_\text{OOD} = 9$) and commercial truck ($n_\text{ID} = 10$, $n_\text{OOD} = 10$) had limited representation; estimates for commercial truck in particular carry high uncertainty and should be interpreted with caution.

\subsubsection{Metrics}
Per-class precision, recall, and F1 score were computed using macro averaging over the target classes. Overall accuracy was computed as the fraction of events for which the predicted label matched the ground-truth label. For each class, the abstention rate was computed as the fraction of ground-truth events of that class for which the pipeline returned \textit{unknown}. All metric computations were performed using scikit-learn \cite{sklearn_api}.

\section{Results}
\label{sec:results}

This section reports classification performance under two evaluation conditions: in-distribution (Ann Arbor N.~Division, two sessions, October 2022) and out-of-distribution (instrumented bicycle open dataset, two trips, July--August 2023). Results are reported for all six vehicle categories. Large van and commercial truck had limited ground-truth representation in both datasets; estimates for commercial truck in particular ($n = 10$ in each condition) carry high uncertainty and should be interpreted accordingly.

\subsection{In-Distribution Performance}
\label{sec:results_id}

The pipeline was evaluated on a pooled set of 3,805 annotated passing events drawn from the two Ann Arbor sessions (1,988 events on October~6; 1,817 events on October~11) across all six vehicle categories.

\subsubsection{Overall accuracy and per-class metrics}
\begin{figure}[h]
    \centering
    \includegraphics[width=\linewidth]{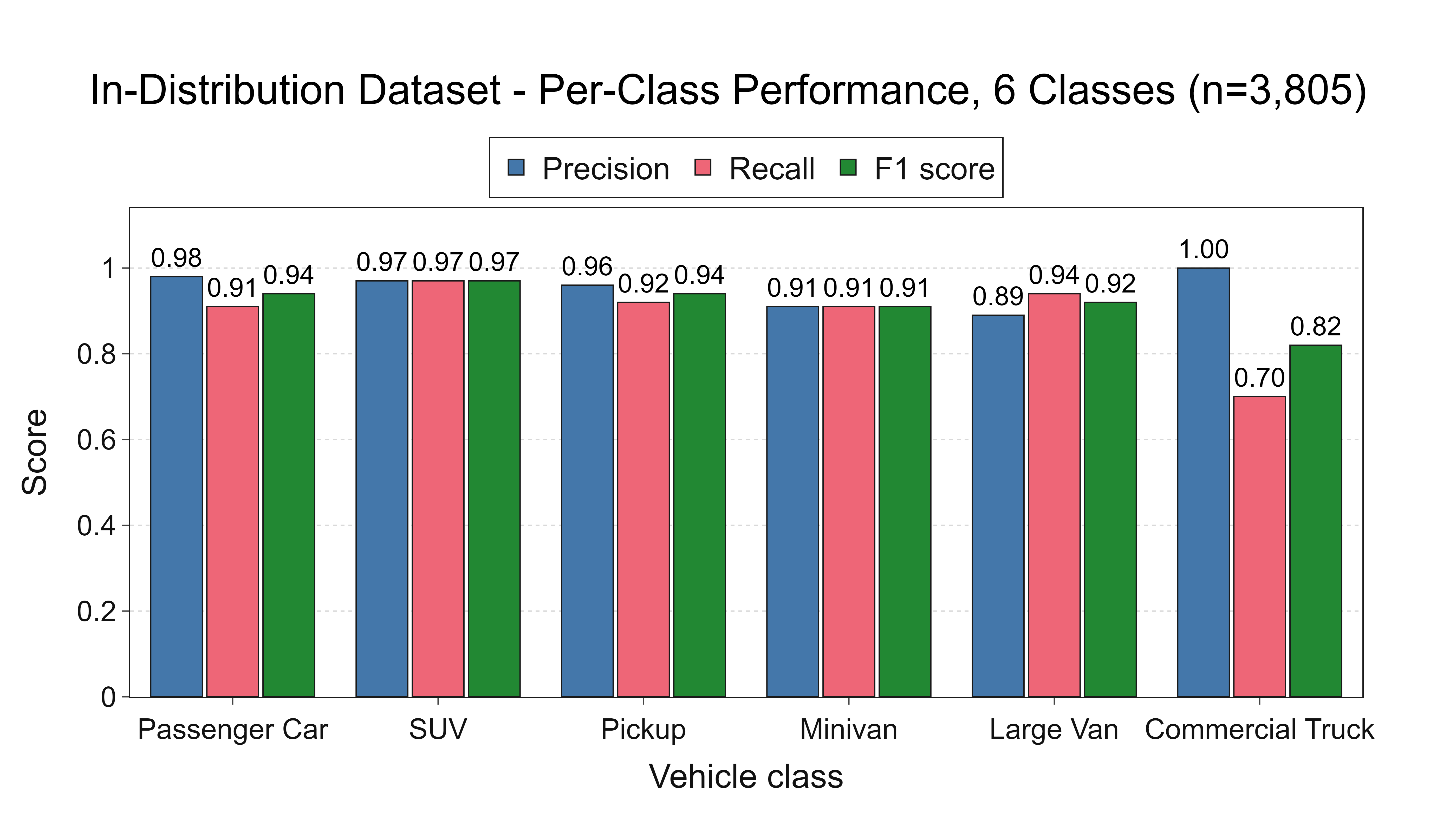}
    \caption{Per-class precision, recall, and F1 score for the in-distribution evaluation set ($n = 3{,}805$). Unknown predictions are counted as incorrect classifications.}
    \label{fig:prf1_id}
\end{figure}

The pipeline achieved an overall accuracy of 0.94. Table~\ref{tab:id_metrics} summarizes per-class precision, recall, and F1 score, and Figure~\ref{fig:prf1_id} illustrates these values.

\begin{table}[h]
\centering
\caption{Per-class classification performance on the in-distribution evaluation set (Ann Arbor N.~Division, pooled, $n = 3{,}805$). Unknown predictions are counted as misclassifications. Commercial truck support ($n = 10$) is too small for reliable estimation.}
\label{tab:id_metrics}
\begin{tabular}{lcccc}
\hline
\textbf{Class} & \textbf{Sample Size} & \textbf{Precision} & \textbf{Recall} & \textbf{F1} \\
\hline
Passenger car     & 1,416 & 0.98 & 0.91 & 0.94\\
SUV               & 1,984 & 0.97 & 0.97 & 0.97\\
Pickup truck      &   217 & 0.96 & 0.92 & 0.94\\
Minivan           &   126 & 0.91 & 0.91 & 0.91\\
Large van         &    52 & 0.89 & 0.94 & 0.92\\
Commercial truck  &    10 & 1.00 & 0.70 & 0.82\\
\hline
\textbf{Overall accuracy} & \multicolumn{3}{c}{\textbf{0.94}} & \textbf{3,805} \\
\hline
\end{tabular}
\end{table}

SUV was the best-classified category (F1 = 0.97). Passenger car had the highest precision (0.98) but a recall of 0.91, reflecting a moderate abstention rate (see Section~\ref{sec:results_id_unk}). Pickup truck and minivan achieved F1 scores of 0.94 and 0.91, respectively. Large van achieved an F1 of 0.92 (precision = 0.89, recall = 0.94) despite having only 52 ground-truth events; the slight precision deficit suggests occasional false positives where a vehicle of another type was assigned the large van label. Commercial truck recorded a perfect precision (1.00) but a recall of 0.70, meaning 3 of the 10 ground-truth events were not correctly classified. Given the sample size, this estimate carries high uncertainty.

\subsubsection{Confusion matrix}
\begin{figure}[h]
    \centering
    \includegraphics[width=\linewidth]{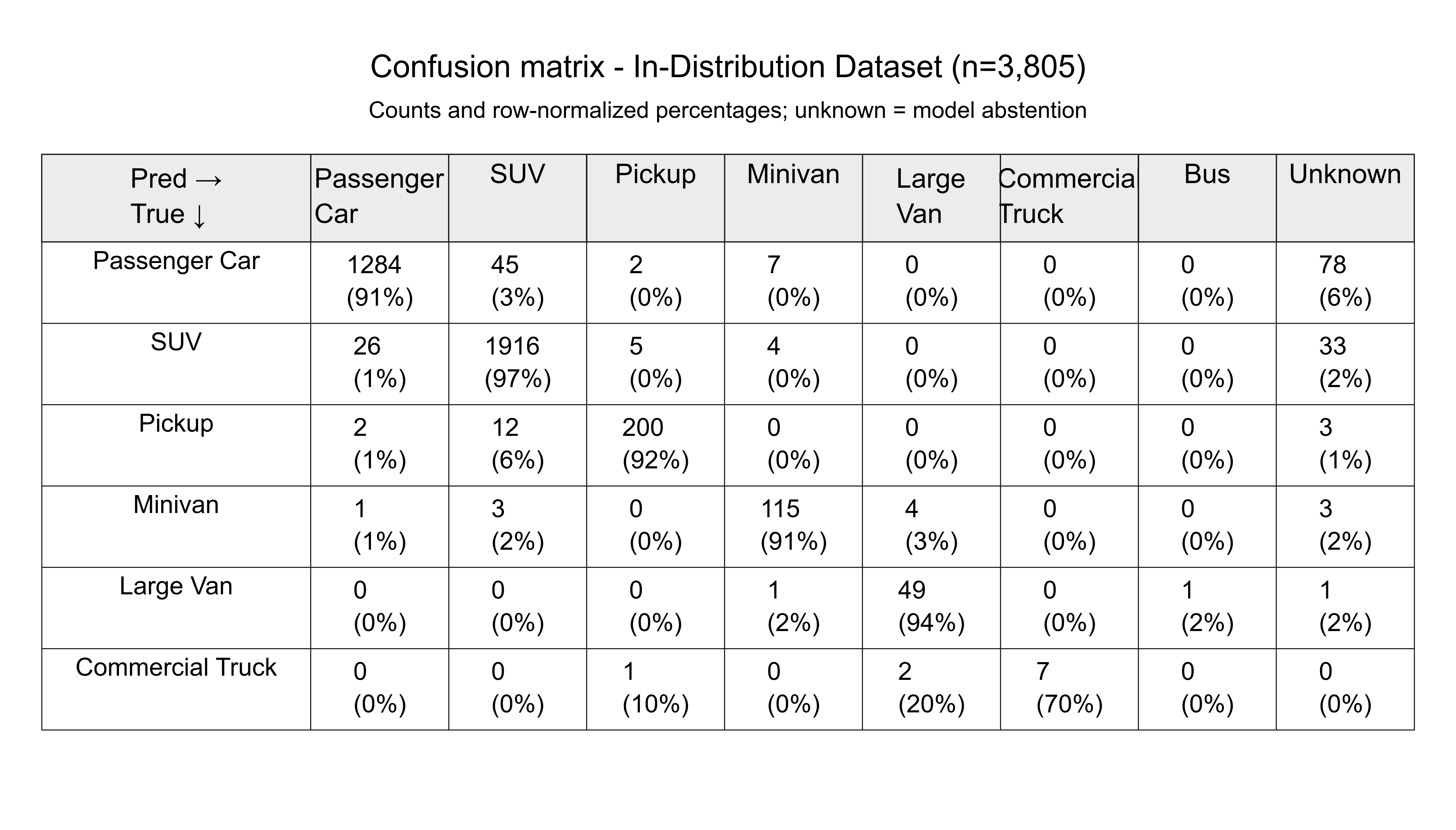}
    \caption{Confusion matrix for the in-distribution evaluation set ($n = 3{,}805$). Each cell shows the count and the row-normalized percentage. The ``unknown'' column represents model abstentions.}
    \label{fig:cm_id}
\end{figure}

Figure~\ref{fig:cm_id} shows the pooled confusion matrix. The dominant off-diagonal errors involve passenger car predicted as SUV ($n = 45$, 3.2\% of passenger car events) and SUV predicted as passenger car ($n = 26$, 1.3\% of SUV events), consistent with the visual similarity between smaller SUVs and sedans. Minivan was occasionally confused with SUV ($n = 3$) and large van ($n = 4$). Pickup truck confusion was minor, with 12 events predicted as SUV and 2 as passenger car.

\subsubsection{Abstention rate}
\label{sec:results_id_unk}
\begin{figure}[h]
    \centering
    \includegraphics[width=0.85\linewidth]{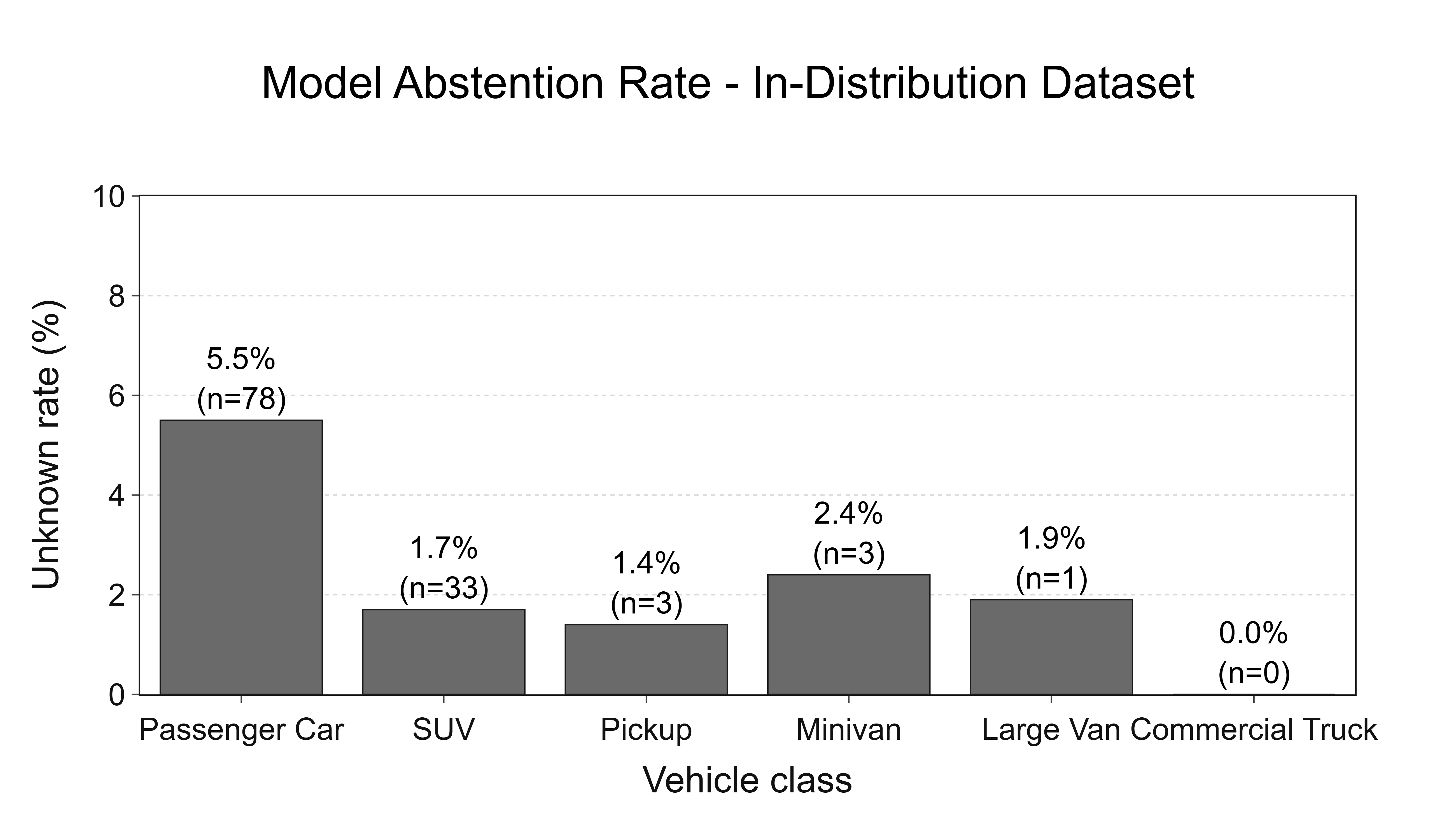}
    \caption{Model abstention rate per class for the in-distribution evaluation set. Values show the percentage of ground-truth events of each class for which the pipeline returned \textit{unknown}.}
    \label{fig:unknown_id}
\end{figure}

Figure~\ref{fig:unknown_id} shows the per-class abstention rate. Passenger car had the highest rate at 5.5\% (78 events), while SUV, pickup truck, minivan, and large van ranged between 1.4\% and 2.4\%. Commercial truck had zero abstentions across its 10 evaluated events. The elevated passenger car abstention may reflect greater intra-class variability across body styles (sedan, hatchback, wagon) relative to the more visually distinct SUV category.

\subsubsection{Qualitative inference examples}
\begin{figure}[h]
    \centering
    \includegraphics[width=\linewidth]{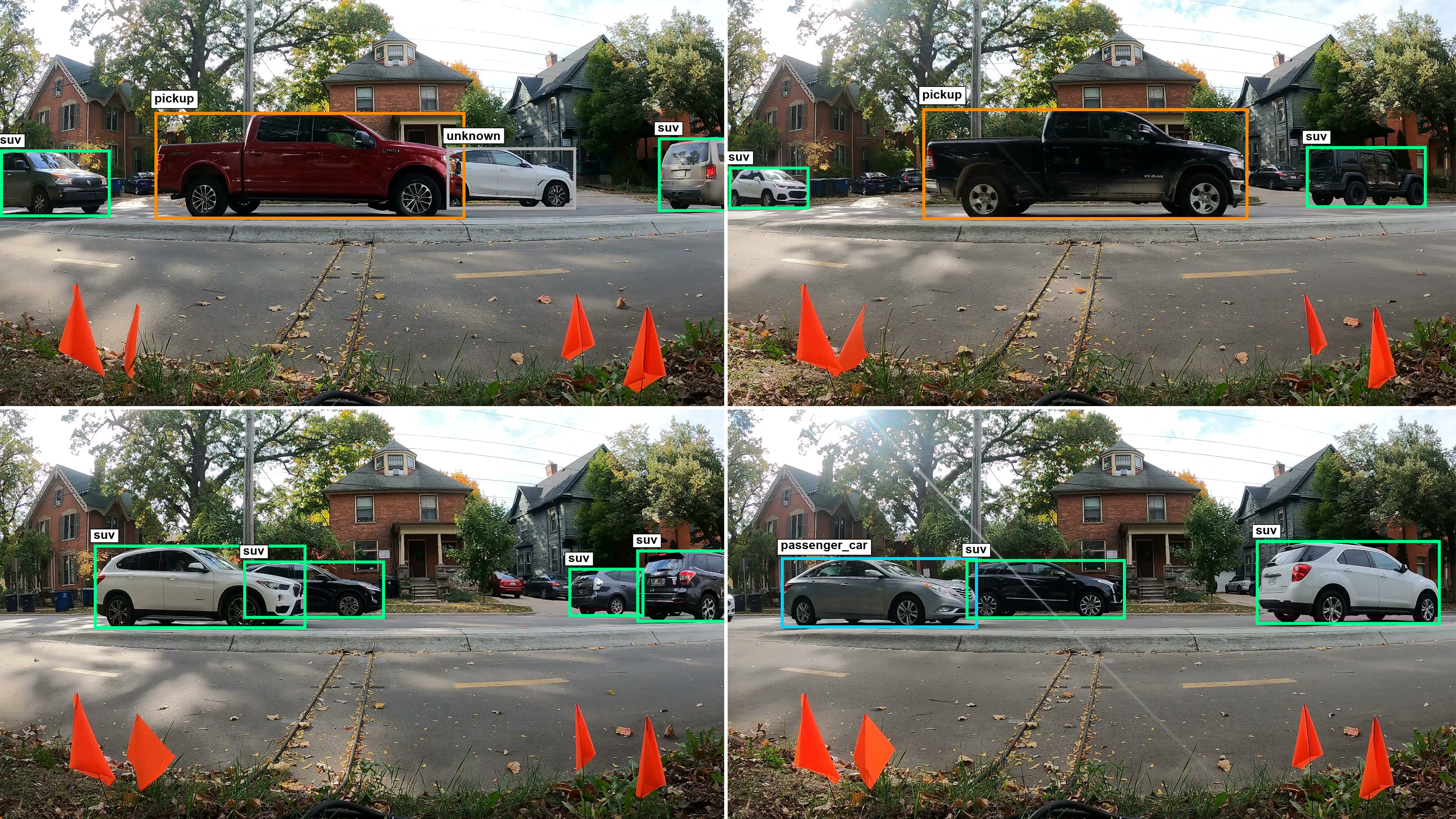}
    \caption{Sample pipeline outputs on in-distribution frames from the Ann Arbor N.~Division dataset. Bounding box colors correspond to predicted vehicle class. The gray \textit{unknown} box (top-left pair) illustrates the confidence-threshold abstention mechanism on a partially visible vehicle.}
    \label{fig:inference_id}
\end{figure}

Figure~\ref{fig:inference_id} shows representative output frames from the Ann Arbor N.~Division dataset. Each frame is processed in all-vehicles mode, where every Stage~1 detection above the confidence threshold is classified and annotated. Color-coded bounding boxes distinguish vehicle types (e.g., green for SUV, orange for pickup truck, cyan for passenger car), and detections for which Stage~2 confidence fell below the threshold are labeled \textit{unknown} in gray. The examples illustrate the pipeline handling multiple simultaneous vehicles, partial occlusions at the frame edge, and the abstention case where a partially visible white vehicle behind the primary overtaking vehicle returns an \textit{unknown} label rather than a potentially erroneous classification.

\subsubsection{Per-session consistency}
The pipeline was applied independently to the October~6 session ($n = 1{,}988$, accuracy = 0.95) and the October~11 session ($n = 1{,}817$, accuracy = 0.93). The per-class F1 values were broadly consistent across sessions (within 0.05 for all six classes), indicating that the pooled result is a reliable summary of in-distribution performance.

\subsection{Out-of-Distribution Performance}
\label{sec:results_ood}

The pipeline was applied to the two instrumented bicycle dataset trips without any retraining or threshold adjustment. These trips were recorded at locations and under conditions distinct from the Ann Arbor training domain, making this a direct test of generalization. The NSF ground-truth annotations contained 9 large van and 10 commercial truck events across both trips combined; these categories are included in the evaluation for completeness but their per-class estimates carry high uncertainty given the limited sample sizes. The combined OOD evaluation set contained 311 annotated events across all six classes after filtering.

\subsubsection{Overall accuracy and per-class metrics}
\begin{figure}[h]
    \centering
    \includegraphics[width=\linewidth]{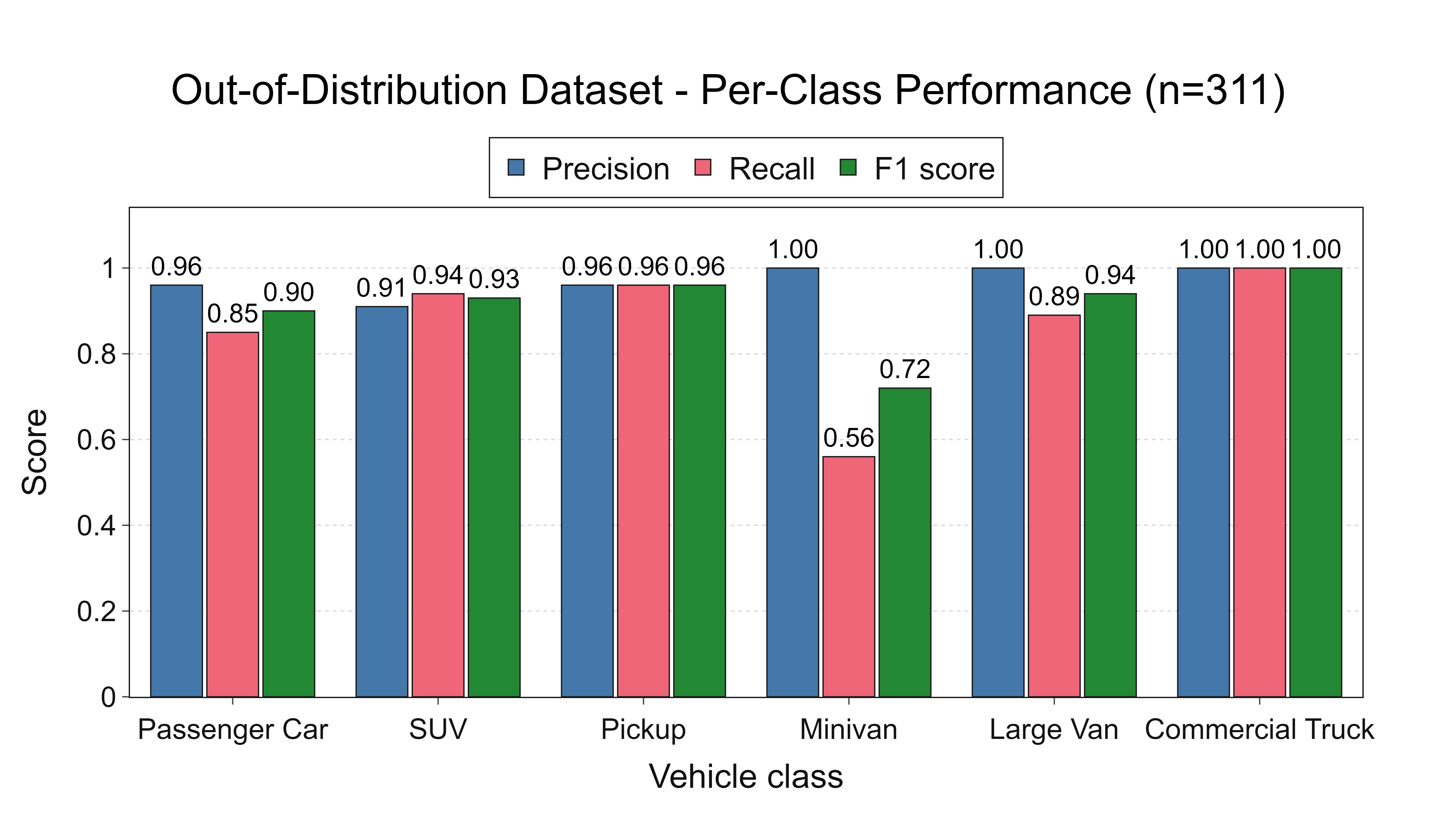}
    \caption{Per-class precision, recall, and F1 score for the out-of-distribution evaluation set (instrumented bicycle open dataset, $n = 311$).}
    \label{fig:prf1_ood}
\end{figure}

Overall accuracy on the OOD set was 0.89, a reduction of 5 percentage points relative to in-distribution performance (0.94). Table~\ref{tab:ood_metrics} provides per-class results, and Figure~\ref{fig:prf1_ood} shows the corresponding bar chart.

\begin{table}[h]
\centering
\caption{Per-class classification performance on the out-of-distribution evaluation set (instrumented bicycle open dataset, pooled trips, $n = 311$). Large van ($n = 9$) and commercial truck ($n = 10$) estimates carry high uncertainty.}
\label{tab:ood_metrics}
\begin{tabular}{lcccc}
\hline
\textbf{Class}  & \textbf{Sample size} & \textbf{Precision} & \textbf{Recall} & \textbf{F1}\\
\hline
Passenger car     & 126 & 0.96 & 0.85 & 0.90\\
SUV               & 123 & 0.91 & 0.94 & 0.93\\
Pickup truck      &  27 & 0.96 & 0.96 & 0.96\\
Minivan           &  16 & 1.00 & 0.56 & 0.72\\
Large van         &   9 & 1.00 & 0.89 & 0.94\\
Commercial truck  &  10 & 1.00 & 1.00 & 1.00\\
\hline
\textbf{Overall accuracy} & \multicolumn{3}{c}{\textbf{0.89}} & \textbf{311} \\
\hline
\end{tabular}
\end{table}

Among the well-represented categories, pickup truck was the most stable class (F1 = 0.96), matching its in-distribution score. Passenger car (F1 = 0.90) and SUV (F1 = 0.93) declined modestly. Minivan showed the largest degradation, with F1 dropping from 0.91 (in-distribution) to 0.72, driven by a recall decline to 0.56; nearly half of the minivan events were either abstained or misclassified. Large van achieved F1 = 0.94 on 9 events and commercial truck F1 = 1.00 on 10 events; while these figures appear strong, the sample sizes are too small to support reliable conclusions about generalization for these categories.

\subsubsection{Confusion matrix}
\begin{figure}[h]
    \centering
    \includegraphics[width=\linewidth]{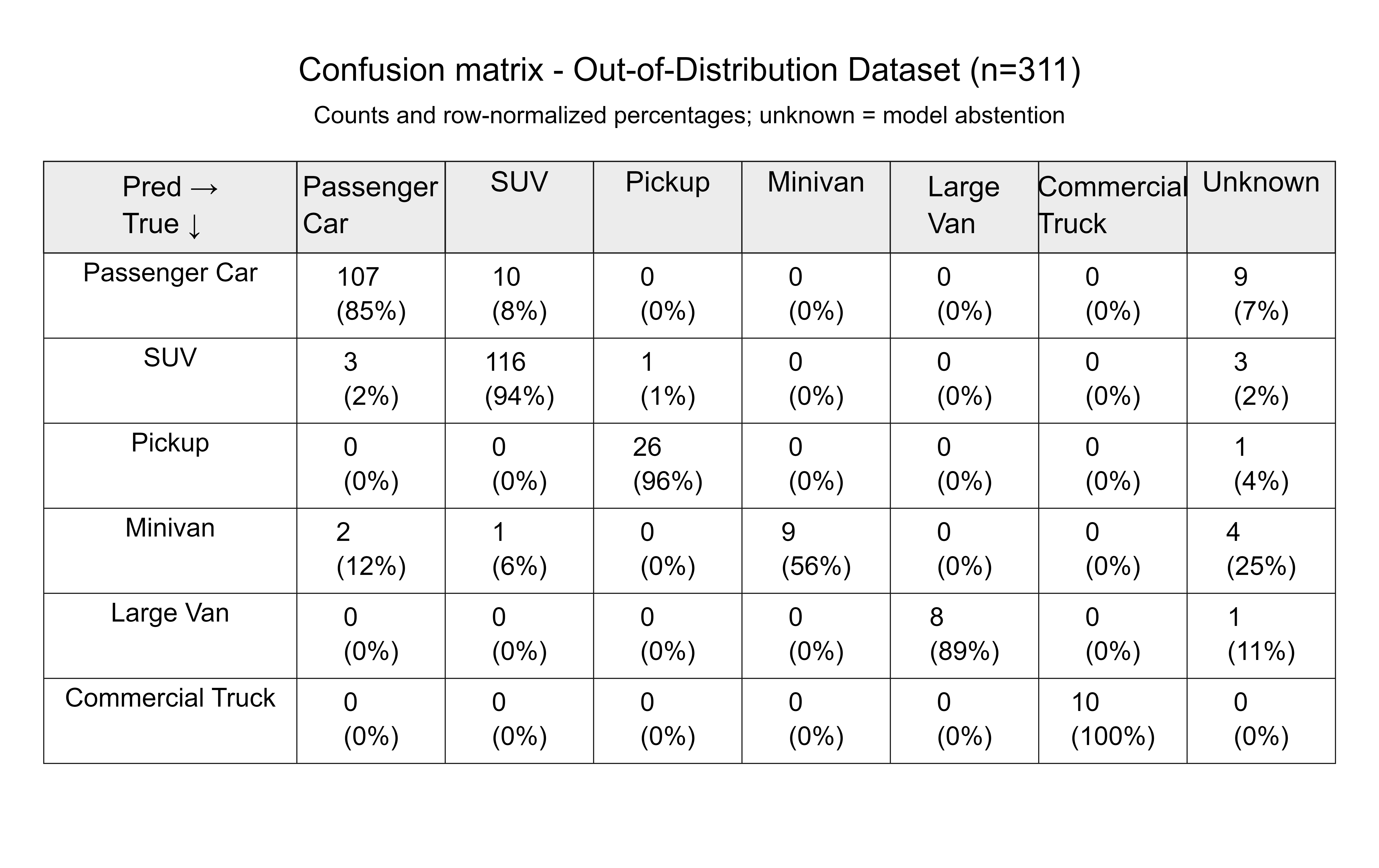}
    \caption{Confusion matrix for the out-of-distribution evaluation set ($n = 311$). Cell format and column conventions follow Figure~\ref{fig:cm_id}.}
    \label{fig:cm_ood}
\end{figure}

Figure~\ref{fig:cm_ood} shows the OOD confusion matrix. The most frequent errors were passenger car misclassified as SUV (10 events, 7.9\%) and SUV misclassified as passenger car (3 events, 2.4\%), mirroring the pattern seen in the in-distribution condition. Minivan was confused with passenger car (2 events), SUV (1 event), and abstained on 4 events, accounting for the low recall for that class.

\subsubsection{Qualitative inference examples}
\begin{figure}[h]
    \centering
    \includegraphics[width=\linewidth]{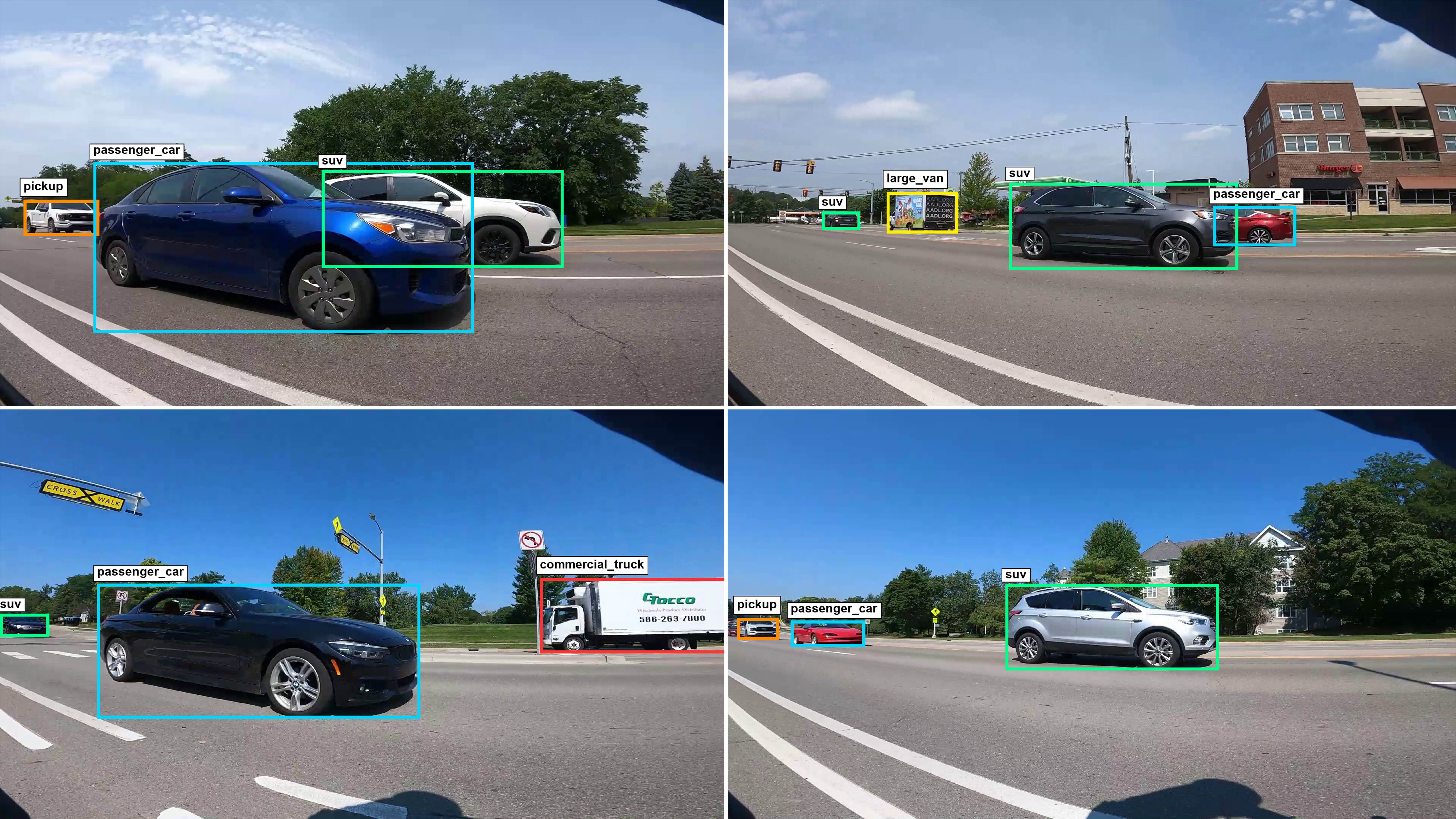}
    \caption{Sample pipeline outputs on out-of-distribution frames from the instrumented bicycle open dataset. The scene geometry, lighting conditions, and vehicle mix differ from the Ann Arbor training domain. The pipeline correctly identifies large van (top-right) and commercial truck (bottom-left) alongside common categories.}
    \label{fig:inference_ood}
\end{figure}

Figure~\ref{fig:inference_ood} shows representative output frames from the instrumented bicycle open dataset. Compared to the Ann Arbor footage, the OOD frames reflect a different camera perspective, road geometry, and broader range of encountered vehicle types. The pipeline correctly identifies the large van and commercial truck in the examples, illustrating that Stage~2 classification generalizes to minority categories under domain shift.

\subsection{Comparative Analysis: In-Distribution vs. Out-of-Distribution}
\label{sec:results_comparison}

\begin{figure}[h]
    \centering
    \includegraphics[width=\linewidth]{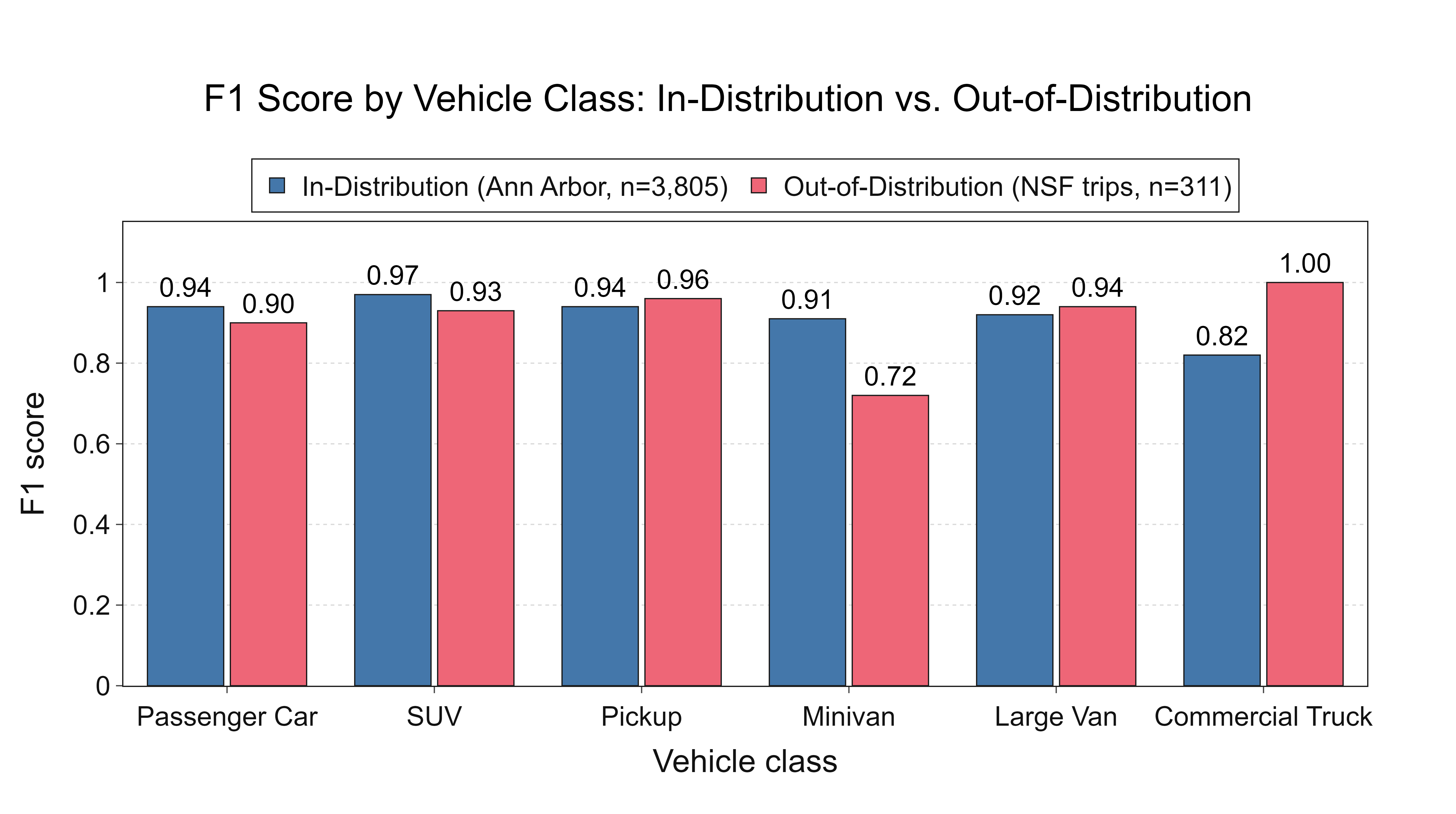}
    \caption{F1 score comparison between in-distribution (Ann Arbor, $n = 3{,}805$) and out-of-distribution (NSF trips, $n = 311$) evaluation sets across all six vehicle classes.}
    \label{fig:f1_comparison}
\end{figure}

Figure~\ref{fig:f1_comparison} places in-distribution and out-of-distribution F1 scores side by side for all six classes. Among the well-represented categories, three maintained F1 at or above 0.90 under the domain shift. The most notable degradation was in minivan (0.91 to 0.72), while passenger car declined by 0.04 points and SUV by 0.04 points. Large van and commercial truck results for the OOD condition reflect very small samples and should be treated as illustrative rather than conclusive.

\begin{figure}[h]
    \centering
    \includegraphics[width=\linewidth]{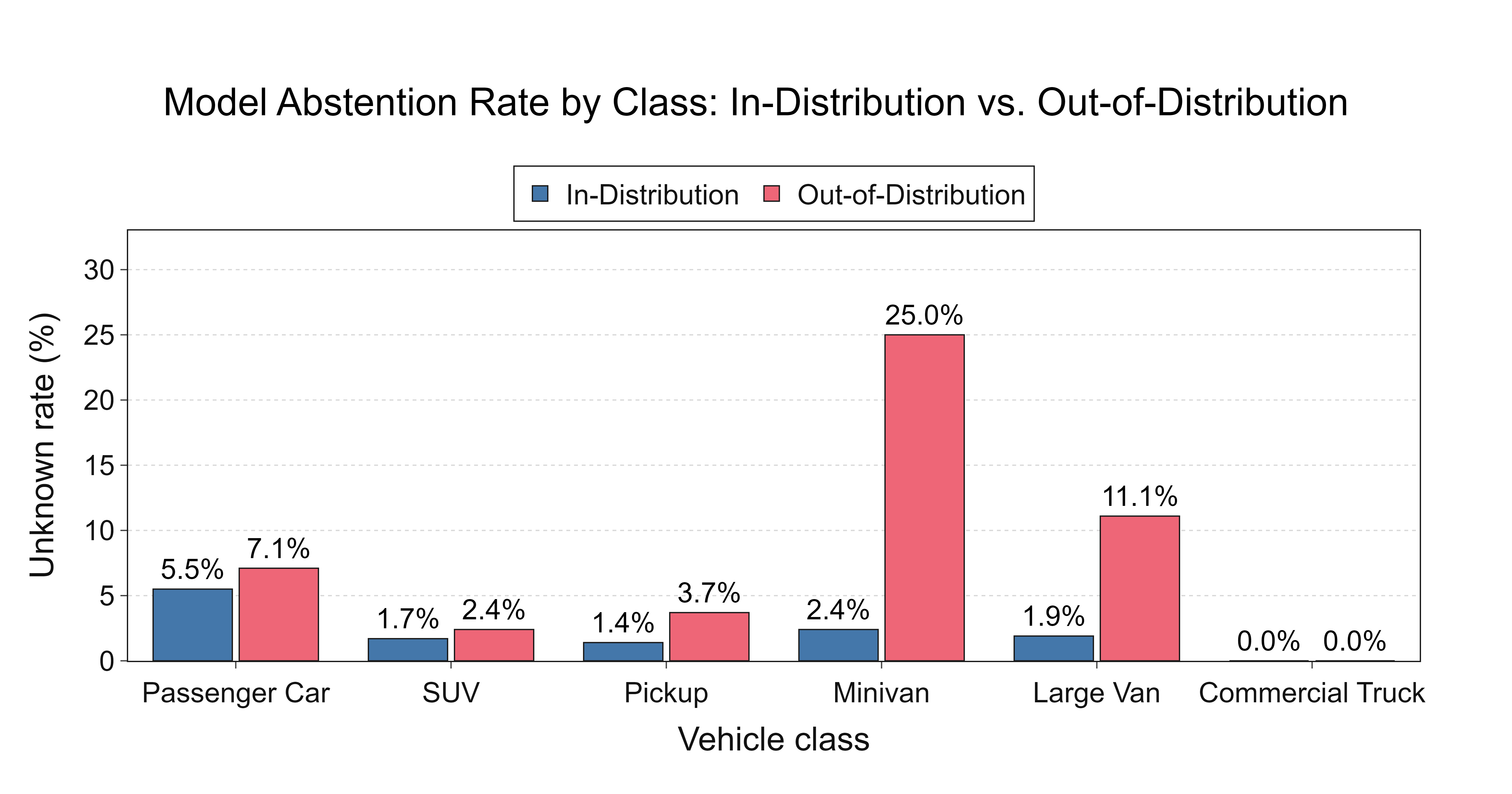}
    \caption{Model abstention rate per class for in-distribution and out-of-distribution conditions. The large abstention rate for minivan in the OOD condition (25\%) accounts for most of the recall gap observed in Table~\ref{tab:ood_metrics}.}
    \label{fig:unknown_comparison}
\end{figure}

Figure~\ref{fig:unknown_comparison} compares abstention rates between the two conditions. Passenger car abstention increased modestly from 5.5\% (in-distribution) to 7.1\% (out-of-distribution). The most pronounced difference was for minivan, where the abstention rate rose from 2.4\% to 25.0\%, consistent with the recall drop observed in Table~\ref{tab:ood_metrics}. This pattern suggests that the confidence-threshold mechanism is correctly reflecting reduced model certainty on out-of-domain minivan examples rather than silently misclassifying them. Large van OOD abstention was 11.1\% (1 of 9 events); commercial truck had no abstentions in either condition, though the sample size limits the interpretability of this observation.

\section{Discussion}
\label{sec:discussion}

\subsection{Interpretation of Pipeline Performance}
\label{sec:disc_performance}

The overall accuracy of 0.94 on the in-distribution evaluation set, achieved without any end-to-end retraining of the detection backbone, demonstrates that a modular coarse-to-fine architecture can yield competitive fine-grained recognition performance in naturalistic roadway conditions. The strength of this result lies not merely in the aggregate figure, but in what the per-class profile reveals about the design choices made. SUV emerged as the best-differentiated category (F1\,=\,0.97), owing to its relatively high training-set representation, visually distinctive silhouette, and the absence of a visually adjacent boundary class at higher specificity. By contrast, passenger car, despite having the largest support among all classes, showed a higher abstention rate (5.5\%) than any other category, which we attribute to the substantial intra-class morphological diversity spanning sedans, hatchbacks, coupes, and wagons. This pattern confirms a well-known challenge in fine-grained recognition: class boundaries defined by functional body type do not always correspond to coherent visual clusters in feature space~\cite{corrales2019cnns,ulkhairi2023dcnn}.

The 5-percentage-point accuracy reduction from the in-distribution to the out-of-distribution
condition (0.94 to 0.89) reflects a modest but non-trivial domain shift attributable to
differences in camera mounting geometry, road geometry, and scene context between the Ann
Arbor corridor and the instrumented bicycle open dataset locations. Domain shift of this magnitude is
consistent with reported generalization gaps in traffic-domain object detection when models
are evaluated across recording sites without explicit adaptation~\cite{wang2025sdgyolov8}. The most instructive pattern is minivan degradation: F1 fell from 0.91 to 0.72, driven almost entirely by a rise in abstention rate from 2.4\% to 25.0\%, rather than by an increase in active misclassification. This behavioral profile suggests that the confidence-threshold mechanism is functioning as intended, propagating genuine model uncertainty to the output rather than masking it with erroneous predictions. That three of the four well-represented classes (SUV, pickup truck, and passenger car) maintained F1\,\(\geq\)\,0.90 under domain shift, without any domain-adaptation retraining, indicates reasonable generalization of the ViT's ImageNet-21k pre-trained representations to roadside imagery.

\subsection{Relationship to Prior Work}
\label{sec:disc_prior_work}

Fine-grained vehicle classification has been approached predominantly through CNN-based pipelines trained on controlled benchmarks. Quintanar et al.\ trained multiple CNN variants on the CompCars dataset and reported top-1 validation accuracy of up to 97.62\% across 431 car models~\cite{corrales2019cnns}; however, such make-and-model benchmarks operate under assumption of clean, frontal, and near-uniform background imagery, which does not reflect the variable viewpoints and outdoor occlusion patterns present in roadway video. Similarly, studies using InceptionNet, ResNet, and VGG for fine-grained vehicle body-type recognition have demonstrated strong accuracy under controlled acquisition conditions but have seldom reported out-of-distribution generalization to camera setups distinct from the training domain~\cite{ulkhairi2023dcnn}. The present work differs by explicitly constructing an out-of-distribution evaluation on independently collected roadway footage, enabling a principled assessment of deployment robustness rather than held-out test-set accuracy alone.

The use of Vision Transformers for traffic-domain classification is relatively nascent. Transformer-based approaches for traffic-sign recognition~\cite{naz2025cvit} and road accident situation detection~\cite{hajri2022vit} have demonstrated that the global self-attention mechanism generalizes well to cluttered road scenes, consistent with the performance observed here. The two-stage architecture adopted in this work follows the broader coarse-to-fine detection-then-classification paradigm common in 3D object detection for autonomous driving~\cite{shi2020pvrcnn} and in dual-stage feature specialization networks for visual perception~\cite{liu2025dsfsn}, which similarly decouple region localization from attribute-level recognition. Distinguishing features of the present system relative to these antecedents include: (i) the deliberate use of a pre-trained, non-fine-tuned RT-DETR backbone for Stage~1, avoiding the need for annotated bounding-box data; (ii) the abstention mechanism at Stage~2, which provides a calibrated confidence signal; and (iii) evaluation on naturalistic overtaking-event data collected from a bicycle-lane corridor, a deployment context not represented in the prior fine-grained vehicle recognition literature.

The truck taxonomy classification system of Almutairi et al.\ \cite{almutairi2022truck} represents the closest precedent in terms of application domain, deploying a CNN trained on highway camera footage to distinguish tractor-trailer configurations for freight flow modeling. The present work extends this line of inquiry to a finer-grained, six-category body-type vocabulary applicable to both freight-relevant categories (commercial truck, large van) and passenger vehicle categories relevant to cyclist exposure research. The deep transfer learning review by Liu et al.\ \cite{liu2023transfer} identifies the core challenge of domain generalization in intelligent vehicle perception as a distributional mismatch between source and target domains; the out-of-distribution evaluation results reported here provide an empirical data point on the magnitude of this gap for roadside ViT-based classifiers without explicit domain-adaptation strategies.

\subsection{Limitations}
\label{sec:disc_limitations}

Several limitations constrain the scope of inference that can be drawn from the current evaluation. First, the training dataset is substantially imbalanced: commercial truck and large van together account for fewer than 200 training samples (approximately 1.1\% of the total), and the evaluation support for these categories is correspondingly sparse (10 and 52 in-distribution events; 10 and 9 out-of-distribution events). While focal loss and weighted random sampling partially mitigate the training imbalance, the uncertainty on per-class estimates for these minority categories is high enough that strong conclusions about their absolute classification performance cannot be drawn. Future studies should prioritize targeted data collection for these categories, particularly at camera sites where commercial trucks are more prevalent such as arterial freight corridors.

Second, the evaluation geometry is specific to overtaking events captured from a fixed roadside camera mounted adjacent to a bicycle lane. The evaluation geometry is specific to fixed roadside cameras mounted adjacent to bicycle lanes; the annotation and matching protocol used here is not directly transferable to other camera configurations such as overhead intersection cameras, vehicle-mounted cameras, or drone footage. Performance on imagery with substantially different viewpoints, including the front-facing or rear-facing perspectives common in dashcam studies, has not been evaluated and cannot be inferred from the current results.

Third, no validation split was held out during Stage 2 training; the model was trained on the full assembled dataset for a fixed 30 epochs without early stopping, which was intentional to maximize exposure to minority classes but means in-training generalization was not monitored. While the separately collected field evaluation sets serve as a proxy for generalization assessment, the absence of a held-out validation curve means that training convergence was assessed by epoch completion rather than by an objective stopping criterion, which could introduce overfitting to the assembled training distribution. Additionally, the training images sourced from Stanford Cars represent manufacturer studio and promotional photography, which differs systematically from the roadside crop imagery encountered at inference; this domain gap within the training set itself may contribute to the elevated abstention rates observed for passenger car.

Fourth, the current pipeline operates on individual frames without temporal aggregation across video. Overtaking events in naturalistic roadway footage typically span several consecutive frames, and a classification system that pools predictions across frames using techniques like majority vote or a Bayesian temporal filter, could reduce abstention rates and improve robustness to frame-level noise such as motion blur and partial occlusion. The single-frame evaluation protocol adopted here represents a conservative baseline that likely understates the performance achievable with temporal smoothing.

\subsection{Implications for Cycling Safety and Beyond}
\label{sec:disc_implications}

The primary motivation for fine-grained vehicle classification in this work is the characterization of overtaking vehicle exposure for cyclists. Epidemiological research on cycling safety has established that crash risk and injury severity are not uniform across vehicle types: large vehicles such as commercial trucks, vans, and SUVs impose disproportionately greater injury risk to cyclists than passenger cars due to differences in frontal geometry, ride height, and impact energy~\cite{monfort2023bicyclist}. Crash databases and police reports typically record only the broad COCO-level category (car, truck) or a coarse administrative category, which conflates passenger SUVs with commercial trucks and conflates minivans with large vans. The six-class vocabulary implemented here maps onto safety-relevant distinctions that are obscured in standard crash reporting, enabling exposure-adjusted analyses of overtaking risk by vehicle body type without requiring manual review of large video archives.

Beyond cycling safety, the pipeline's open-source release and demonstrated generalization to an independent recording site suggests its utility as infrastructure for a range of transportation research tasks. Automated vehicle-type counts disaggregated to body type enable traffic composition monitoring without the dedicated roadside sensors (inductive loops, pneumatic tubes, and weigh-in-motion platforms) currently required for vehicle classification in traffic counts. Planners using passive video data from existing infrastructure cameras could apply the pipeline to assess whether local vehicle fleets are shifting toward larger body types (SUV penetration, commercial van growth), which has implications for urban road design, parking geometry, and pedestrian sight-line modeling. In the autonomous vehicle perception domain, the results contribute a field-validated benchmark for transformer-based fine-grained recognition under the naturalistic occlusion and lighting conditions of roadside deployment, which complements the predominantly synthetic or controlled-setting evaluations that dominate the AV perception literature~\cite{liu2023transfer}.

The confidence-based abstention mechanism has specific relevance for safety-critical applications. Rather than emitting a potentially erroneous label with high apparent certainty, the pipeline degrades gracefully by returning \textit{unknown} when the softmax distribution is insufficiently concentrated. In the context of an automated traffic monitoring system, \textit{unknown} events can be routed to human review or flagged for re-analysis with an alternative model, preserving data quality without discarding the event entirely.

\subsection{Future Research Directions}
\label{sec:disc_future}

Several extensions would meaningfully advance the capabilities demonstrated here. First, temporal aggregation across consecutive frames through majority voting, a recurrent classifier, or a transformer operating over frame sequences, would leverage the redundancy inherent in video data to reduce per-event error and abstention rates, particularly for visually ambiguous categories such as minivan and passenger car. Second, domain adaptation strategies, including adversarial feature alignment~\cite{liu2023transfer} or test-time adaptation using batch statistics from the target site, could narrow the in-distribution to out-of-distribution accuracy gap without requiring site-specific labeled data. Third, expanding the training set for commercial truck and large van through targeted field collection at freight-intensive corridors, augmented by synthetic rendering of these categories from 3D CAD models, would stabilize per-class estimates for the minority categories that are most relevant to cyclist injury risk. Fourth, extending the taxonomy to include trailer type, vehicle color, and approximate size category would increase the richness of the exposure record and enable finer-grained risk stratification in epidemiological studies. Finally, integration of the pipeline with trajectory estimation by leveraging the bounding-box time series to infer passing distance and relative speed, would close the gap between vehicle-type classification and actionable cyclist safety metrics, including the lateral clearance at overtaking events targeted by passing-distance legislation \cite{feng2018drivers} in many jurisdictions.

\section{Conclusions}
\label{sec:conclusions}

This paper presented a two-stage, open-source computer vision pipeline for fine-grained vehicle body-type classification from naturalistic roadway video. The system combines a pre-trained RT-DETR detector for coarse vehicle localization with a fine-tuned Vision Transformer (ViT-Base/16) for six-category classification (passenger car, SUV, pickup truck, minivan, large van, and commercial truck), and incorporates a confidence-based abstention mechanism that withholds predictions when the Stage~2 softmax output falls below a class-specific threshold of 0.60.

Evaluated on 3,805 annotated overtaking events from a shared bicycle-lane corridor in Ann Arbor, Michigan, the pipeline achieved an overall accuracy of 0.94, with per-class F1 scores ranging from 0.91 for minivan to 0.97 for SUV. Per-session results were consistent across two independent recording dates (October~6 accuracy = 0.95; October~11 accuracy = 0.93), indicating that the pooled performance estimate is stable and not attributable to conditions specific to a single collection session. The dominant classification errors involved confusion between visually adjacent categories, specifically passenger car and SUV, which accounted for the majority of off-diagonal entries in the confusion matrix and reflect the known challenge of distinguishing body types that lack sharp visual discontinuities in roadside imagery.

On an independent out-of-distribution evaluation set of 311 events drawn from the instrumented bicycle open dataset, collected at separate sites and under distinct recording conditions, the pipeline achieved an overall accuracy of 0.89 without any retraining or threshold adjustment. Three of the four well-represented categories (SUV, pickup truck, and passenger car) maintained F1 scores at or above 0.90 under this domain shift, demonstrating reasonable generalization of the ViT's ImageNet-21k pre-trained representations to roadside imagery from unseen locations. The most substantial cross-domain degradation was observed for minivan (F1 declining from 0.91 to 0.72), driven primarily by a rise in the abstention rate from 2.4\% to 25.0\% rather than by an increase in active misclassification. This pattern reflects the abstention mechanism functioning as intended: propagating genuine model uncertainty to the output label rather than producing silent errors.

The six-category vocabulary implemented in this pipeline maps directly onto body-type distinctions that are safety-relevant but absent from standard crash databases and coarse COCO-level detection outputs. Large vehicles, including SUVs, pickup trucks, and commercial vans, are disproportionately associated with severe cyclist injuries, yet this granularity is not captured in conventional traffic monitoring or crash records. By enabling automated, scalable annotation of vehicle body type from existing roadside video, the pipeline supports exposure analyses that can quantify the composition of
overtaking traffic relative to crash records~\cite{hamann2017beyondgps}, evaluate the behavioral effects of passing-distance legislation~\cite{feng2018drivers}, and inform road design decisions. The open-source release of the full pipeline, including model weights, training code, inference scripts, and ground-truth annotation files, is intended to lower the barrier to replication and adaptation across new recording sites and research applications.

Several limitations bound the scope of the present findings. Per-class estimates for commercial truck and large van carry high uncertainty due to sparse evaluation support (10 and 52 in-distribution events, respectively), and conclusions about their generalization performance should be treated with caution. The evaluation geometry is specific to fixed roadside cameras mounted adjacent to
bicycle lanes; adaptation to other viewpoints (overhead intersection cameras, dashcams, or drone footage) would require adapting the annotation and matching protocol. Additionally, the pipeline classifies individual frames independently without temporal aggregation, which likely understates the performance achievable by pooling predictions across the multiple frames that typically constitute a single overtaking event.

In summary, the results demonstrate that a modular coarse-to-fine transformer architecture, combining a general-purpose detector with a fine-tuned vision transformer, can achieve robust fine-grained vehicle classification on naturalistic roadway video without requiring bounding-box annotation for training and without site-specific retraining for deployment at new locations. The system provides a practical and reproducible tool for vehicle-type exposure measurement in cycling safety research, traffic composition monitoring, and related transportation analysis applications.

\section{Code and Data Availability}
\label{sec:availability}

The complete pipeline implementation, including Stage~1 and Stage~2 inference scripts, training code, evaluation utilities, and the fine-tuned ViT model weights, is publicly available at: \url{https://github.com/[repo]}. The repository also includes the ground-truth annotation files for the Ann Arbor N.~Division evaluation sessions and instructions for reproducing all reported results. The out-of-distribution evaluation data were drawn from the instrumented bicycle open dataset, accessible at 
\url{https://fenggroup.org/facilities/lidar-bike.html#open-data-repository}.

\section{Acknowledgments}
\begin{itemize}
    \item This work is supported by the National Science Foundation under award number 2142757.
    \item Grammarly was occasionally used for grammar and spelling check.
\end{itemize}

\bibliographystyle{plain} 
\bibliography{ref} 

\end{document}